\documentclass[10pt,journal,compsoc]{IEEEtran}
\usepackage{graphicx}
\usepackage{tabularx,booktabs}
\usepackage{multirow}
\usepackage{amsmath}
\usepackage{booktabs}
\usepackage{array}
\usepackage{float}
\usepackage[linesnumbered,ruled,vlined]{algorithm2e}
\ifCLASSOPTIONcompsoc
\usepackage[nocompress]{cite}
\else
\usepackage{cite}
\fi
\begin{document}
\title{EILearn: Learning Incrementally Using Previous Knowledge Obtained From an Ensemble of Classifiers}

\author{Shivang Agarwal\IEEEauthorrefmark{1}, 
        C. Ravindranath Chowdary\IEEEauthorrefmark{1}, Shripriya Maheshwari\IEEEauthorrefmark{2}
        
\IEEEcompsocitemizethanks{\IEEEcompsocthanksitem \IEEEauthorrefmark{1} Department
of Computer Science and Engineering, IIT (BHU), Varanasi-221005.\protect\\
E-mail: shivanga.rs.cse16@itbhu.ac.in, rchowdary.cse@itbhu.ac.in
\IEEEcompsocthanksitem \IEEEauthorrefmark{2} IIITDM Jabalpur\protect\\
E-mail:2016254@iiitdmj.ac.in 
}

}

\IEEEtitleabstractindextext{
\begin{abstract}
We propose an algorithm for incremental learning of classifiers. The proposed method enables an ensemble of classifiers to learn incrementally by accommodating new training data. We use an effective mechanism to overcome the stability-plasticity dilemma. In incremental learning, the general convention is to use only the knowledge acquired in the previous phase but not the previously seen data. We follow this convention by retaining the previously acquired knowledge which is relevant and using it along with the current data. The performance of each classifier is monitored to eliminate the poorly performing classifiers in the subsequent phases. Experimental results show that the proposed approach outperforms the existing incremental learning approaches.

\end{abstract}
\begin{IEEEkeywords}
Ordering sentences, Text processing, Structuring document 
\end{IEEEkeywords}}

\maketitle

\IEEEdisplaynontitleabstractindextext

\IEEEpeerreviewmaketitle

\ifCLASSOPTIONcompsoc
\IEEEraisesectionheading{\section{Introduction}\label{sec:introduction}}
\else
\section{Introduction}
\label{sec:introduction}
\fi
\IEEEPARstart{T}he major challenge in incremental learning is to learn from new data without accessing the previously observed data. This property fosters the stability-plasticity dilemma, where stability describes retaining the previously acquired knowledge, and plasticity describes learning knowledge from new data. So, an ideal approach for incremental learning must find a balance between stability and plasticity. Focussing only on plasticity may lead to a situation, called \textit{catastrophic forgetting}. At the same time, concentrating only on stability may lead to loss of knowledge. Only the relevant previous knowledge must be preserved. Irrelevant previous knowledge must be discarded, but with the ability that it can be recalled whenever required. 

Another challenge in incremental learning is learning in the presence of concept drift. Concept drift is a situation where the underlying data distribution changes over time, such that $p_{t+1}(x, w) \neq p_{t}(x, w)$. Here, $x$ represents an instance, and $w$ represents the class label associated with $x$.

Existing ensemble-of-classifiers based models are successful in learning incrementally by passing the query instances and assigning weights dynamically to the classifiers. The final class is assigned to each instance after taking the weighted average of all the classifiers.

We propose EILearn, an Efficient Incremental Learning approach, which does not rely on ensemble learning to generate weak classifiers. Instead, we make use of a clustering algorithm to form clusters of records. Each of these clusters is used to generate a hypothesis. These hypotheses are used as an ensemble to decide the class label for a query instance. The decision is taken using majority voting and base rating methods explained in the subsequent sections. 

\section{Related Work}
Most of the work on incremental learning uses ensemble learning to generate weak classifiers that can be used to decide the class label for the query instance. Learn++ is the family of algorithms designed for incremental learning, each of which is capable of handling a different issue, such as concept drift, imbalanced data distribution, etc. \cite{983933} \cite{1007781} \cite{10.1007/3-540-44938-8_18}. An incremental SVM learning approach is proposed in \cite{Laskov:2006:ISV:1248547.1248616}, which has substantial storage requirements because it stores the support vectors in the memory during the entire learning period. In this paper, we propose a novel approach where we do not have to rely on ensemble learning to produce an ensemble of classifiers. We do that by making use of a clustering algorithm, which forms clusters of records. Our approach is different in its architecture discussed in Section 3. To the best of our knowledge, this architecture has never been used for any incremental learning approach. 

\section{Methodology}
We start with taking a dataset and dividing it into two parts, one for training and testing, and other for validation. The data for training and testing $T$ is split up into $k$ parts to demonstrate the $k$ phases of incremental learning. The validation data $V$ remains same for all the phases from $T_1$ to $T_k$. Later, each $T_i$ is partitioned into the separate train and test sets, $P_i$ and $Q_i$ respectively. In every phase, $P_i$ is used to perform clustering on it and form clusters of records. These clusters are used to generate hypotheses. Each cluster generates one hypothesis, the accuracy of which is tested on corresponding $Q_i$. 

Therefore, we get a number of hypotheses at the end of every phase. These hypotheses are used as an ensemble of classifiers, which classifies a query instance by majority voting. Each of these hypotheses is associated with a base rating. For every correct classification, the base rating of the hypothesis is increased. These base ratings of hypotheses are useful in case of a tie, where we get the equal number of votes for more than one classes. In that case, we consider the decision made by the hypothesis with the highest rating. 

After the end of one phase, we are able to calculate the average of individual accuracies of classifiers on test data $Q_i$, as well as on validation data $V$. In addition to this, we also calculate the accuracy of the ensemble of classifiers on test data $Q_i$ as well as on validation data V. 

We use the same methodology in the subsequent phases. The only difference is, in the later phases we have to import those hypotheses generated in the previous phase(s) that have accuracy more than 50\%, the reason being it is incremental learning; in every phase, we have to learn something new while retaining the previous relevant knowledge. The hypotheses that have accuracy lower than 50\% are discarded and kept in a buffer so that any of them can be recalled when all the current hypotheses misclassify the query instance. Now, when we test the accuracy of the ensemble of classifiers which also includes the hypotheses from the previous phase(s), it is expected to increase because of increased knowledge. The incremental behavior of our model is reflected in the enhanced accuracy of the group of hypotheses on the validation data when we move to the subsequent phases of learning.

\begin{algorithm}[!t]
 \KwData{Dataset $D$ with $n$ instances;
 
 Number of phases $k$;
 
%
 A buffer $B$ for discarded hypotheses;
 
 Base learner $L$;
 
 Clustering algorithm $C$;}
 \KwResult{Hypotheses $H_k$ generated in each of the $k$ phases;}
 
 Split the dataset into two parts $T$ and $V$; $T$ is for train \&test, and $V$ is for validation;
 
 Partition $T$ in $k$ parts $T_1$ to $T_k$;
 
 \For {$i$= 1 to $k$}
 {Partition $T_i$ into train \& test sets, $P_i$ \& $Q_i$;
 
 $C_i \leftarrow C(P_i)$; \tcp{To generate $m$ clusters}
 
 $L_i \leftarrow C_i$;
 
 $H_i\leftarrow \sum_{j=1}^{m}L_i$;
 
 Test the $H_i$ on $Q_i$;
 
 $B \leftarrow H_q, where\ Accuracy(H_q)<50\%$;

 Test $H_i$ on $V$;
 
 $H \leftarrow H_i$;
 
 Test the $H$ on $Q_i$ using majority voting;
 
 If none of the hypotheses can classify the current instance, call a hypothesis from buffer $B$ and add it to the current ensemble of hypotheses;
 
 Test the $H$ on $V$ using majority voting;
 
 Handle the case where Tie occurs using base rating method;
 
 Test $H$ in $(k-1)th$ phase on $Q_{k-1}$;
 
 Export the hypotheses generated so far to the subsequent phase;
 }
 \caption{Algorithm EILearn}
\label{Algo1}
\end{algorithm}

\section{Experimental Results}
To test the working mechanism of our proposed approach, we perform experiments on two benchmark datasets. Diabetes dataset is available on UCI repository\footnote{https://archive.ics.uci.edu/ml/datasets/Diabetes}. It has 768 instances and 20 attributes. We partition the dataset into two parts such that, set $T$ has 400 instances and set $V$ consists of 368 instances. Later, we divide set $T$ into four subparts to demonstrate 4 phases of incremental learning. Each phase has 100 instances, out of which 66 are used for training and rest are used for testing. 

\begin{table*}[]
 \centering
 \caption{EILearn results on Diabetes dataset}\label{Result1}
 \begin{tabular}{|c|c|c|c|c|c|}
 \hline
    \textbf{Dataset/Incremental Training} & \textbf{Average/Learner} & \textbf{Training1}& \textbf{Training2} & \textbf{Training3}& \textbf{Training4} \\
     \hline     
 $Q_1$ & 53.53\% & 45.45\% &  &  & \\ 
 \hline
 $Q_2$ & 65.14\% &  & 69.69\% &  & \\
  \hline
 $Q_3$ & 65.65\% &  &  & 66.66\% & \\
  \hline
 $Q_4$ & 67.67\% &  &  &  & 72.72\%\\
  \hline
 \textbf{Validation} &  \textbf{61.58\%} &  \textbf{33.15\%} & \textbf{67.39\%} &  \textbf{70.10\%} &  \textbf{77.98\%}\\
  \hline
 \end{tabular}
 \end{table*}
 
Experimental results are described in Table \ref{Result1}. In our experiments, we use EM algorithm \cite{Celeux:1992:CEA:146597.146608} for clustering and J48 \cite{Quinlan1993} as the base classifier. The choice of clustering algorithm and base classifier can be varied depending on the nature of the problem. In the column "Average/Hypothesis," we show the average accuracies of individual hypotheses on the corresponding test datasets $Q_i$. In the subsequent column, we show the accuracies of the ensemble of hypotheses over test data in the current phase. The last row of the Table \ref{Result1} shows the accuracies tested over validation data. The second column in the last row shows the average of all individual accuracies of hypotheses tested on validation data. The following cells in the last row show the increased accuracy of the ensemble of hypotheses during the learning phases. It is evident that the accuracy increases significantly as we move from phase 1 to phase 4. 

 \begin{table*}[]
 \centering
 \caption{EILearn results on King Rook VS King Pawn dataset}\label{Result2}
 \begin{tabular}{|c|c|c|c|c|c|}
 \hline
    \textbf{Dataset/Incremental Training } & \textbf{Average/Learner} & \textbf{Training1}& \textbf{Training2} & \textbf{Training3}& \textbf{Training4} \\
     \hline     
 $Q_1$ & 60.60\% & 56.28\% &  &  & \\ 
 \hline
 $Q_2$ & 58.47\% &  & 67.06\% &  & \\
  \hline
 $Q_3$ & 70.21\% &  &  & 80.23\% & \\
  \hline
 $Q_4$ & 70.45\% &  &  &  & 86.22\%\\
  \hline
 \textbf{Validation} &  \textbf{64.79\%} &  \textbf{51.83\%} &  \textbf{63.62\%} &  \textbf{86.62\%} &  \textbf{92.47\%}\\
  \hline
 \end{tabular}
 \end{table*}
Our second dataset is King Rook VS King Pawn, available on UCI repository\footnote{https://archive.ics.uci.edu/ml/datasets/Chess+\%28King-Rook+vs.+King-Pawn\%29}. It contains 3196 instances. We partition the dataset into two parts $T$ and $V$. Set $T$ has 2000 instances and set $V$ has 1196 instances. Later, we divide $T$ into four parts to demonstrate 4 phases of incremental learning. In each phase, we will have 500 instances, where 66.6\% of those are used for training and rest are used for testing. 

Table \ref{Result2} shows the experimental results performed over King Rook VS King Pawn dataset. Similar to the experiments conducted on Diabetes datasets, here also the accuracy of the ensemble of hypotheses on the validation data increases as we move from learning phase 1 to phase 4, which shows the incremental behavior of our model. 

\section{Conclusions}
We proposed EILearn, an efficient incremental learning approach. The working mechanism is explained to show the difference between existing incremental learning approaches and the proposed approach. Our model is different in its architecture. Instead of relying on ensemble learning methods, it makes use of clustering to generate multiple classifiers. We use majority voting along with base rating method to decide the class label for the query instance. Following the proposed model, we are able to get good experimental results over real-world datasets. 

\bibliographystyle{IEEEtran}

\end{document}